# Explainable AI applications in the Medical Domain: a systematic review

Nicoletta Prentzas, Antonis Kakas, and Constantinos S. Pattichis

*Abstract*—Artificial Intelligence in Medicine has made significant progress with emerging applications in medical imaging, patient care, and other areas. While these applications have proven successful in retrospective studies, very few of them were applied in practice. The field of Medical AI faces various challenges, in terms of building user trust, complying with regulations, using data ethically. Explainable AI (XAI) aims to enable humans understand AI and trust its results. This paper presents a literature review on the recent developments of XAI solutions for medical decision support, based on a representative sample of 198 articles published in recent years. The systematic synthesis of the relevant articles resulted in several findings: (1) model-agnostic XAI techniques were mostly employed in these solutions (2), deep learning models are utilized more than other types of machine learning models, (3) explainability was applied to promote trust, but very few works reported the physician's participation in the loop, (4) visual and interactive user interface is more useful in understanding the explanation and the recommendation of the system. More research is needed in collaboration between medical and AI experts, that could guide the development of suitable frameworks for the design, implementation, and evaluation of XAI solutions in medicine.

*Index Terms*—Explainable Artificial Intelligence (XAI), Explainable Machine Learning, Medical Decision Support Systems, Medical XAI.

Part of this work was undertaken under the University of Cyprus internal project, Integrated Explainable AI (IXAI) for Medical Decision Support, ARGEML 8037P-22046. This work is also partly funded by the project 'Atherorisk' "Identification of unstable carotid plaques associated with symptoms using ultrasonic image analysis and plaque motion analysis", code: Excellence/0421/0292, funded by the Research and Innovation Foundation, the Republic of Cyprus.

The third author acknowledges partial funding from the CYENS Centre of Excellence via the European Union's Horizon 2020 research and innovation program under grant agreement No. 739578 and the funding received from the Government of the Republic of Cyprus through the Deputy Ministry of Research, Innovation and Digital Policy.

## I. Introduction

BROADLY defined as the imitation of human cognition by a machine, recent interest in Artificial Intelligence (AI) has been driven by advances in machine learning (ML), in which computer algorithms learn from data without human direction [1]. These advances in ML algorithms, together with increased computing power and the exponential growth of data, have made the application of AI in practical fields possible such as in content recommendation, machine translation, and facial recognition. AI has also found its way into more critical domains such as healthcare, autonomous driving, and criminal justice [2].

The advantages of AI and ML in healthcare have been extensively discussed in the literature [1], [3]. A ML based system is typically trained with data to recognise patterns or associations, to learn how to automatically perform a task (e.g. classification) on a new set of inputs. These self-learning capabilities can be used to detect abnormalities in medical images, predict a diagnosis or recommend a treatment. Studies [4], [5] already suggest that AI can perform comparatively well with humans at key healthcare tasks, for example in lung cancer screening [6] (radiology), or the automatic prediction of malignant colonic lesions [7] (gastroenterology). Yet, the current adoption of AI in medicine suggests a significant but unfulfilled opportunity, as the Medical AI community battles with the various challenges for their safe and effective deployment. Challenges arise as a result of the AI algorithms opacity, and their inability to explain their results in a way that humans can understand [8]. Take for example an algorithm that accurately detects breast cancer in medical images. Physicians may be able to verify the outcome of the algorithm, but, do they have sufficient reasons to trust the diagnosis when they cannot determine how it was derived? What if the diagnosis is wrong and the physician cannot verify the result? Concerns about accountability, responsibility, and trust emerge with opaque AI. Moreover, legal and privacy aspects, e.g. with the European General Data Protection Regulation (GDPR), make opaque approaches difficult to use, as their inability to explain why a decision has been made does not allow for the decision to be contested. Explainability in AI, as the ability of the system to explain its results, has become a critical requirement [9].

Explanations for medical decision support in real-world scenarios can allow systems to act as peer companions to clinicians and other medical practitioners. For example, a system predicting stroke (or not) from ultrasonic carotid plaques using texture and clinical features would be valuable only if the predictions can be explained to radiologists and cardiologists. A hybrid XAI approach to produce such a system was studied in [10]. This study is based on a general methodology for combining statistical and symbolic learning methods to produce an interpretable model from which explanations are easily extracted. The resulting explainable model is a high-level symbolic theory of argumentation, which generates predictions with explanations in terms of arguments that support the prediction in various ways. Explanations have the general form: "The prediction of stroke/not-stroke is *supported by* reasons A. These reasons are *strengthened against* reasons B supporting the opposite predictions due to reasons C hold." Hence the explanation not only explains why a prediction is supported but it also explains why the opposite prediction might be weaker. This second, *contrastive* element, of explanations is equally important as the first, *attributive* part, of explanations. Such *peer explanations* can help a domain expert understand the line of reasoning of the particular model, verify the model's function and output, or give meaningful feedback in order to improve and refine the model if needed. In addition to that, depending on the role of the user, more informative explanations can be derived, including more reasons to support a prediction, e.g. for regulatory requirement. Explanations therefore act as a tool to justify the decisions of AI systems and contribute in building a high-level of trust towards these systems.

In AI, there was always a strong connection to explainability. In the early days of AI, the predominant reasoning methods were logical and symbolic, by performing some form of logical inference on human-readable symbols [8]. Early medical expert systems like MYCIN [11], NEOMYCIN [12], and CASNET [13] are examples of these approaches. They were successful in terms of performance, but their daily practice was very limited. The majority of these expert systems failed to be accepted by the practitioners because they didn't provide acceptable explanations [14]. Nevertheless, these early AI systems generated a trace of their inference steps, which then became the basis for explanation [15]. However, these systems were not effective. Effectiveness was improved significantly with the introduction of probabilistic learning and the development of machine learning techniques like Support Vector Machines (SVMs), Random Forests, and Deep Neural Networks (DNN). These gave more effective models, which however were opaque and less explainable.

Explainability is attracting much interest in medicine and other domains, with numerous surveys [16–22] demonstrating a large amount of work that is difficult to analyze and synthesize into a broad view of the field. Some of these works provide holistic views of the XAI landscape [16], [17], [23], [24], others review explainability techniques on deep learning [20], [23], [26] or machine learning in general [19], [27],[28], agents and robots [23], [29], or in a specific domain of applications like in healthcare [9], [28], [30] and the medical sector [29], in banking [31], and in the industry [32]. The contribution of this work is to provide an overview of the current state of research in the area of explainable AI/ML for medical decision support (not restricted to any type of medical data), both from a technical and medical-application point of view. While our study has similarities with related works in the literature, we emphasized on the explainability from the physician's view as the domain expert and the recipient of the explanation. Therefore, the scope of this work mainly concerns the technical aspects of the proposed solutions (baseline models, XAI techniques, data types), as well as their explainability features (type, scope, presentation) and how these addresses the challenges of XAI from the perspective of the physician, to understand and trust AI in medical practice. While more use cases of the explanation can be realized in a medical AI system, involving different user roles (e.g. developer, regulator), and different goals (auditability, fairness, safety, ethical use of data, and more), these are out of the scope of this work.

The paper is organized as follows: in Section II we present the methodology of this study, the literature selection strategy, and the data extraction process. In Section III we provide definitions, discuss concepts and challenges, and in Section IV we suggest a taxonomy of XAI techniques and provide basic information. In Section V we present the results of the literature review selection process and their assessment. Finally, we present our conclusions and perspectives of future work in Section VI.

## II. METHODOLOGY

We start with an introduction to the main concepts relating to explainable AI, discuss explainability challenges, in general and specifically in the medical domain, and suggest a taxonomy of XAI techniques that can be applied to new problems. We also provide basic information and references to commonly used techniques. Then, we use this taxonomy to categorize the articles selected, and further assess their explainability characteristics, in an attempt to understand the impact of recent XAI developments in medicine. We also evaluated a list of representative studies, that reported a human-in-the-loop approach, using explainability assessment metrics. We defined a set of questions to help us discsuss the findings of this review: "Which XAI techniques have been used and for which medical use case?", "Is there a connection between the XAI techniques and the medical use case or the medical data the solution has been applied to?", "Is the explanation useful, has it been evaluated by the physician?", "Can we apply these solutions in clinical practice?", "To what extent are the explainability challenges addressed by the proposed solutions and which areas need further development?".

### A. Literature review strategy

We carried out a literature review based on the Preferred Reporting Items for Systematic Review and Meta-Analyses (PRISMA) framework [33]. While we did not aim at exhaustive literature search, to find a sufficient number of articles, three major digital databases were selected and searched: (a) Scopus, which provides access to journals from the life, social, physical and health sciences; (b) IEEE Xplore digital library, which provides access to various engineering and/or technology related resources, and, (c) PubMed, which provides access to biomedical literature from MEDLINE and life sciences journals. We searched for relevant articles published in journals at any time until October 2022. Although the terms interpretability and explainability are used interchangeably in the literature, the term "Explainable AI (XAI)" is primarily associated with this evolving area of research and we therefore decided to focus this review on explainability. At first, we run a search query in Scopus (search#1, Table I) which returned 2285 documents. Then, we revised this search query using medical terms (search#2, Table I) which returned 720 documents. At this point, it is worth mentioning that ~30% of explainable AI/ML research results is in the medical domain, and this observation justifies the general purpose of this review. Also, noteworthy is the exponential increase in publications over the last three years. In particular, the number of research results for 2022 is nearly equal to the total of previous years. For that reason, we decided to narrow down the results of 2022 to papers with high impact (in terms of citations and / or the resource journal). Thus, the records identified using Scopus was reduced to 405. Similarly, we applied the search#2 query in the PubMed and IEEE databases which returned 303 (from 421) and 51 documents respectively. The total number of records identified through search queries was 759 in which we added 10 from other resources. After removing duplicates (292) we screened titles and abstracts of the remaining 477 studies based on a set of inclusion and exclusion criteria:

- Inclusion criteria
  (i) Explainability is part of the problem described in the abstract and the proposed solution is trying to address. (ii) The solution proposed was applied to a specific medical-related task (a diagnostic task, risk prediction, or medical decision support). (iii) The explainability technique is part of the "explainable modeling" or "post-modeling" process in the machine learning pipeline (see Fig. 3).
- Exclusion criteria
  (i) The subject of the article was not in the medical or other relevant domain (genetics, bioinformatics, etc), (ii) Review/ Survey/ Opinion articles, and, (iii) General XAI topic/ algorithms (not applied/tested on medical data and/or to support a medical decision).

TABLE I
LITERATURE SEARCH QUERIES

| Search#1 | Results: Scopus =2285 |
|---|---|
| query: [TITLE=explainab* OR Keywords=explainab*] AND [ABSTRACT contains "machine learning" OR "deep learning" OR "artificial intelligence" OR "network" OR "regression" OR "decision tree" OR "random forest" OR "gradient boosting" OR "support vector machine" OR "Naïve Bayes" OR "fuzzy" OR "rules"] | |
| Search#2 | Results: Scopus =405, Pubmed =303, IEEE=51 |
| query: [TITLE=explainab* OR Keywords=explainab*] AND [ABSTRACT contains "machine learning" OR "deep learning" OR "artificial intelligence" OR "network" OR "regression" OR "decision tree" OR "random forest" OR "gradient boosting" OR "support vector machine" OR "Naïve Bayes" OR "fuzzy" OR "rules"] AND [ABSTRACT contains "disease" OR "diagnosis" OR "medical" OR "patient" OR "health*"] | |

The screening process (title and abstract) resulted in the exclusion of 253 studies as unrelated (exclusion reason: other domain or survey/review/other studies or XAI but not in the medical domain) while assessing the remaining 224 articles for their eligibility resulted in the exclusion of 26 more (exclusion reasons: full-text not available or XAI not addressed). In the end, 198 articles[i] were selected for the systematic scoping review. The PRISMA flowchart is illustrated in Fig. 1.

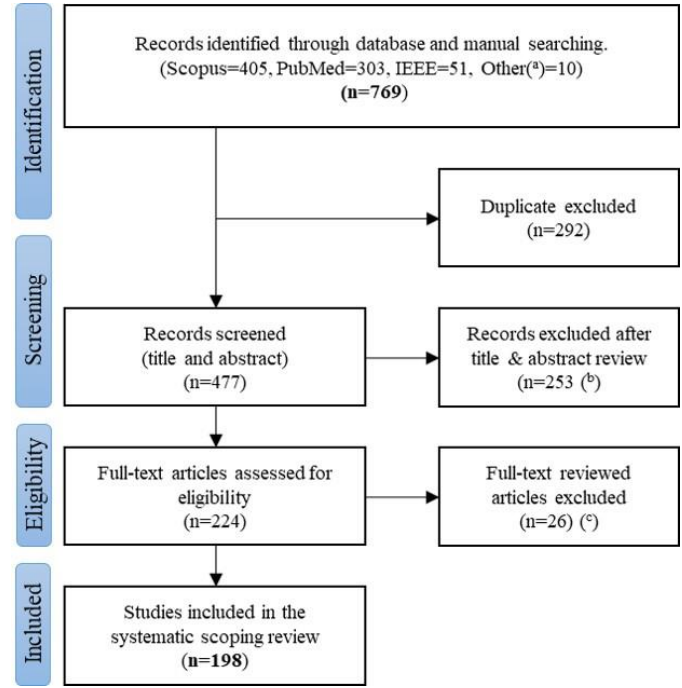

Fig. 1. PRISMA flow diagram. Explainable AI/ML applications in the medical domain. Study selection for Review. ([a]) Selected studies from a previous literature review. ([b]) Exclusion reason: survey/review or XAI-theory or subject not relevant to a medical diagnosis task (adverse drug events, policy development, survey data analysis, medical education, coding of clinical notes, genome-wide association studies). ([c]) Exclusion reason: full-text was not available or XAI was not addressed.

*B. Data extraction*

The literature screening process resulted in the selection of 198 articles related to the scope of this review for full-text evaluation. First, we performed a high-level assessment considering technical characteristics related to the ML baseline model of the proposed solution, the XAI technique employed, and the disease related to the medical data (e.g. cancer). Then, we selected a number of representative studies, that reported a human-in-the-loop approach, to examine their explainability features using available assessment metrics. The results of the aforementioned analysis are presented in Section V as A. "Medical XAI studies overview", B. "Medical XAI studies by XAI technique", and C. "Medical XAI studies, explainability evaluation".

III. XAI CONCEPTS AND CHALLENGES

Before we proceed with our literature review, it is appropriate to start with a common point of understanding about what the term *explainability* is in the context of AI and particularly in ML. In this section, we examine several definitions and concepts in relation to explainable AI as well as the challenges that need to be addressed towards this direction. For this, we will be guided primarily from the perspective of the need for AI application for medical decision support. A summary of these concepts is presented in Fig. 2.

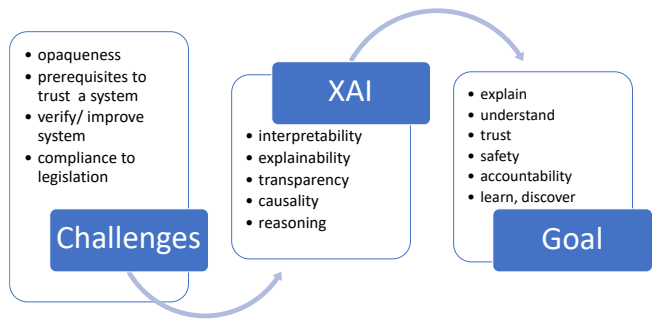

Fig. 2. XAI Related Concepts Summary

*A. Definitions, related concepts*

The problem of explainability is as old as AI and may be the result of AI itself. The term explainability was introduced in 2004 by Van Lent *et al*. [34], in their publication about Full Spectrum Command (FSC), a training system developed for the U.S. Army to achieve a targeted training objective, which included an XAI feature. Today, the term explainability in the literature is associated with the initiatives and efforts made in dealing with the black-box / opaque nature of AI/ML systems, their lack of transparency, and trust concerns [35]. Other relevant terms include understandability [36], transparency [37],[38], causality [15], [39], [40], contestability [41] and justifiability [42].

Samek *et al*. [43] in their effort to explain the need for Explainable AI, referred to social and technical aspects in favor of explainability in AI systems. The ability to explain the rationale behind one's decision is often a prerequisite for establishing a *trust* relationship between people, e.g., when a doctor explains the therapy decisions to her/his patient. Therefore, for explanations to be meaningful, humans need to have a basic understanding of the machine's learning and reasoning processes [44]. Technical or software engineering aspects need to be blended with social aspects. For example, system verification and learning/discovery, need to be combined with conformance to legislation (i.e. GDPR and "the right to explanation" [45]).

An explanation is closely related to the concept of *interpretability*: systems are interpretable if their operations can be understood by a human, either through introspection or through a produced explanation [37]. Doran *et al*. [35] suggested that *reasoning* is a critical step in formulating an explanation about why or how some event has occurred. Truly Explainable AI should formulate, for the user, a line of computation of the output from the input data, using human-understandable features.

Lipton *et al*. [46] suggested that interpretability is not a monolithic concept, it reflects several distinct ideas: it can be described in terms of *transparency* of the machine learning system i.e., the algorithm, features, parameters, and the resulting model should be comprehensible by the end-user. Interpretability may also mean different things for different people and in different use cases [47].

For the purpose of this survey, we define Explainable AI (XAI) as AI/ML in which predictions/results are accompanied by explanations, presented in a way that humans can understand and formulate a line of reasoning that explains/justifies the decision-making process of the model.

*B. Challenges in implementing XAI*

One of the major challenges facing explainable AI is in designing models that generate accurate predictions and understandable explanations. Moreover, for humans to trust a model, information on the causes of their decisions is needed. To build trust, we also need to embed transparency, fairness, and robustness into the AI system. Completeness is also important, meaning that the system should be able to defend its actions, provide relevant responses to questions, and be audited. Further, while a variety of techniques towards XAI have been proposed, evaluating the quality of the explanation, as well as the explainable AI/ML itself, is vital for the application in practice (e.g. in clinical practice). Finally, since this study is focused on the medical domain, it is appropriate to consider unique challenges, if any, that apply in this particular domain. Key challenges are analyzed further in the next paragraphs.

*1) Accuracy vs interpretability trade-off*

Historically, there has been a trade-off between interpretable machine learning models and the performance (precision, recall, F-Score, AUC, etc.) of the prediction models [27]. Inherently and intrinsically interpretable models like decision trees often perform less well on prediction tasks compared to less interpretable or opaque models like deep learning and gradient boosting. Researchers have recently proposed new models which exhibit high performance as well as interpretability. Examples are GA$^2$M [49], rule-based models [50] and model distillation [51].

*2) Trustworthy and Ethical AI*

In software development, trust is built through mechanisms such as testability, auditability, documentation, and many other elements that help establish the reputation of a piece of software. In traditional software development, the behavior of a system is dictated by explicit rules expressed in the code. In the case of AI systems, their behavior is based on the knowledge that evolves over time. Researchers from IBM proposed a new methodology for establishing trust in AI systems [52], whilst the European Union (EU) published a set of guidelines to be taken into consideration during the design of explainable AI systems [53]. IBM's fundamental pillars and EU guidelines are summarized in Table II.

*3) Quality of explanations*

As humans, we start to build trust by asking questions. Questions that experts in AI ask a system are aimed to provide inside explanations, focused on debugging, reliability and validation. These explanations often cannot explain why a system produces a prediction/decision in a way that is precise (true to the model) and understandable to humans. To build trust with technical and non-technical end-users, Leilani *et al*. [19] suggested the development of outside explanations that are interpretable (understandable to humans), complete (true to the model), and answer why questions. However, the most accurate explanations are not easily interpretable to people; and conversely, the most interpretable descriptions often do not provide predictive power. The challenge facing XAI is in the generation of explanations that are both interpretable and complete.

In [54] the authors defined a list of properties for the explanations that can be used to evaluate their quality. Some of these properties are **Accuracy:** is related to the completeness explained in the previous paragraph. **Fidelity**: is associated with how well the explanation approximates the prediction of a model – this property is relevant to the model-induction techniques (explained in the next section). **Stability**: it represents how similar the explanations are for similar instances. **Comprehensibility**: one of the most important properties but at the same time difficult to define and measure. It is related to how well humans understand the explanations. **Importance**: it is associated with how well the explanation reflects the importance of features or of parts of the prediction. **Representativeness**: it describes how many instances are covered by the explanation.

TABLE II
TRUSTWORTHY AND ETHICAL AI

| Term | Definition |
| --- | --- |
| *Diversity, non-discrimination, and fairness* | - AI systems should use training data and models that are free of bias, to avoid unfair treatment of certain groups.<br>- AI systems should consider the whole range of human abilities, skills and requirements, and ensure accessibility. |
| *Robustness, Safety* | - AI systems should be safe and secure, not vulnerable to tampering or compromising the data they are trained on.<br>- Trustworthy AI require algorithms to be secure, reliable and robust enough to deal with errors or inconsistencies during all life cycle phases of AI systems. |
| *Explainability* | - AI systems should provide decisions or suggestions that can be understood by their users and developers. |
| *Lineage* | - AI systems should include details of their development, deployment, and maintenance so they can be audited throughout their lifecycle. |
| *Transparency* | - The traceability of AI systems should be ensured. |
| *Accountability* | - Mechanisms should be put in place to ensure responsibility and accountability for AI systems and their outcomes. |

*4)* Human friendly explanations

The need for human friendly is a general challenge for computing relating to the ease of human understanding of the systems. This is sometimes called Perspicuous Computing [55]. Alan Cooper [56] argues that software is often poorly designed (from a user perspective) because programmers design for themselves, rather than their target audience; a phenomenon he refers to as the 'inmates running the asylum'. Likewise, most of the existing work in explainable AI uses solely the researchers' intuition of what constitutes an appropriate explanation for humans. Explainable AI is more likely to succeed if researchers understand, adopt, implement, and improve models from the vast and valuable bodies of research in philosophy, psychology, and cognitive science [38]. In his in-depth survey about this topic Miller provides a list of human-friendly characteristics of explanations. Major findings include: (i) Explanations are contrastive: people usually don't only ask why a certain prediction was made but rather why this prediction was made instead of another prediction; (ii) Explanations are selective and focus on one or two possible causes and not all causes for the recommendation; (iii) Explanations are part of social interaction between the explainer and the explainee. This means that the social context determines the content, the communication, and the nature of the explanations.

*5)* XAI Frameworks, Design & Evaluation

A framework for the systematic assessment of explainable AI approaches was proposed in [57] to facilitate a common ground for their evaluation and comparison. They suggested a fact sheet to accompany every explainability method designed to assess, its (1) functional and (2) operational requirements, (3) the quality of explanations against a list of usability criteria, (4) security, privacy, and vulnerabilities, and its (5) validation via experiments.

A different, conceptual framework was proposed in [58] for building human-centered decision-theory-driven XAI based on theoretical foundations of human decision making from the fields of philosophy and psychology. Likewise, the authors in [59] discussed key concepts of measurement from a user perspective and suggested methods for evaluating the explanation goodness, the degree of users-in-context satisfaction by explanations, and other user-centered metrics for explainable ai systems. A multidisciplinary and extended survey on explainable ai design and evaluation methods [120] proposed a design and evaluation framework that connects design goals and evaluation methods for end-to-end XAI systems design.

Research effort is currently limited on methods for evaluating the explanation, as the output/result of an explainability technique. Evaluation approaches and metrics discussed in the literature [60],[61] can be broadly grouped into (a) *human-centered* (or user-based or human-grounded) approaches, and, (b) *objective* (or heuristic or functionality-grounded) approaches. Human-centered approaches involve experiments with users to assess the effectiveness of explanations, by means of interviews or questionnaires. Objective evaluations refer to metrics and automated approaches to evaluate methods of explainability. Research shows that current efforts mainly focus on developing metrics for assessing notions of explainability (challenges, goals, properties) or else requirements for the design and implementation of XAI systems. These metrics mostly concerned the assessment of specific XAI requirements, and some of these were discussed in the context of particular explanation types. A summary of these works is shown in Table III.

*6)* Unique challenges in the medical domain

The application of Medical AI in clinical practice faces many challenges, due to their opaqueness and lack of trustworthiness, and related concerns about potential bias, accountability and responsibility. Medical AI systems need to be transparent, understandable, and explainable to gain the trust of physicians, regulators, and patients [62]. But transparency may not always be necessary, as authors in [63] argue that the reliability of algorithms, what they call computational reliabilism, can provide reasons for trusting the outcomes of Medical AI. Besides, even when these outcomes are trustworthy, ethical concerns remain regarding clinical data interpretation, professional responsibility and expertise.

TABLE III
EXPLAINABILITY METRICS

| XAI Goals, Properties | | Assessment Metrics |
|---|---|---|
| **Human-centered evaluations** | | |
| Goal | Understandability Efficiency | **Subjective**: usefulness, satisfaction, confidence, trust. **Objective**: behavioral, physiological and other task-performance metrics. |
| **Objective evaluations** | | |
| | | **Qualitative:** |
| Goal | Understandability Efficiency | Form of cognitive chunks, number of cognitive chunks, compositionally, monotonicity and other interactions between units, uncertainty and stochasticity. |
| | | **Quantitative:** |
| Properties | Accuracy | Correctness |
| | Fidelity | Sensitivity, identify, separability, correctness |
| | Consistency | Implementation invariance |
| | Stability | Stability |
| | Comprehensibility | Compactness |
| | Certainty | |
| | Importance | Sensitivity |
| | Novelty | |
| | Representativeness | Completeness |

While trust is linked to the ability to explain expert recommendations using causal knowledge in their domain of expertise, medicine is a domain in which the underlying causal system is in its infancy; the pathophysiology of the disease is often uncertain, and the mechanisms through which interventions work is either not known or not well understood [64]. At the same time, experts sometimes are not able to provide an explanation because of the various heterogeneous and vast sources of different information [15]. Therefore, a medical AI system in which decisions reflect causal knowledge of the domain that can also be expressed in terms that are accessible to non-experts can help foster trust. The importance of causality in explanation is extensively discussed in [40] by T. Miller where he argues that causal attribution is not alone by itself an explanation. While an explanation refers to causes to explain an event, an event may have many causes, yet only a subset relevant to the context is of interest to the end user, meaning that explanations need to be contextual.

Medical AI poses challenges to developers, medical professionals, and legislators as it requires a reconsideration of roles and responsibilities [42]. For example, from a legal perspective, two core fields for explainability are (a) informed consent, where the underlying process and algorithms have to be explained to the patient in advance, and, (b) the certification and approval as a medical device, with the U.S. Food and Drug Administration (FDA) [65] and the European Commission (MDR) [66] having to introduce requirements to regulate the need for explainability for AI-based medical devices. Towards this direction, FDA recently released the first AI/ML-based software as medical device (SaMD) Action Plan in response to stakeholders' feedback about the proposed regulatory framework [65], [67].

The analysis of representative XAI articles in the medical domain in which physicians participated in the design and/or evaluation of the system, emphasizes the need for an active role of the end user in the development cycle of XAI solutions [68–70]. This is also discussed in the literature as interactive machine learning or human-in-the-loop [71]. For the physician, the explanation is just another piece of information that needs to be examined and verified. Ideally, explanations should be given to the user through a visual and interactive interface, in a simple and comprehensive way, providing all information needed to support the decision process. They call these ideal systems "conversation partners" [68] or "expert companions" [69].

In summary, explainability aims to overcome the barriers arising from the opacity of AI algorithms that relate to technical, ethical, legal and medical issues. Explainable AI challenges from a technical perspective include aspects like testability, auditability, being able to understand a model's internal function to debug and fix errors. From a medical perspective, challenges relate to potential bias, accountability and responsibility, and legal challenges like the informed consent, certification and approval as medical devices, and liability. Explain to gain trust is the ultimate goal, but it has different meanings depending on the recipient of the explanation. Explanations and their presentation, from a user interface perspective, should be tailored to the different user roles to meet their specific requirements, like for example the physicians need for "conversation partners". Guidelines for Trustworthy and Ethical AI, and theoretical frameworks for XAI approach assessments were proposed, that can be utilized in the development of Medical XAI solutions. Finally, quality and human-friendly properties have been suggested, for defining explainability requirements and assessment metrics to evaluate the results.

## IV. XAI TECHNIQUES TAXONOMY

In general, explainability can be applied to the entire ML/AI development life cycle. As illustrated in Fig. 3 with the standard machine learning pipeline, it can be applied (a) at the data preparation phase (*pre-modeling*), (b) during the model building and training (*explainable modeling*), or after the model is trained (*post-modeling*). Designing new or modified learning algorithms for XAI systems, can lead to models that will be able to generate explanations along with their predictions. In this study, we will concentrate mainly on explainable and post-modeling techniques. While the concept of explainability reflects several different dimensions, we have focused on efforts to make the AI system more transparent, so as to briefly present major explainability techniques and taxonomies from the literature. The list of methods presented is not exclusive or exhaustive.

### A. XAI Taxonomy

A number of taxonomies can be found in the literature for the classification of explainability techniques, types of explanation, properties of good explanations, and goals of explanation [8], [16], [19], [22], [26], [37], [62]. We suggest a taxonomy of XAI techniques, as illustrated in Fig. 4 with broad categories that cover the majority of the techniques studied in the articles selected for this review. The proposed taxonomy is used in Section V for the analysis of the studies selected for this review. Table IV lists typical explainability techniques from the literature.

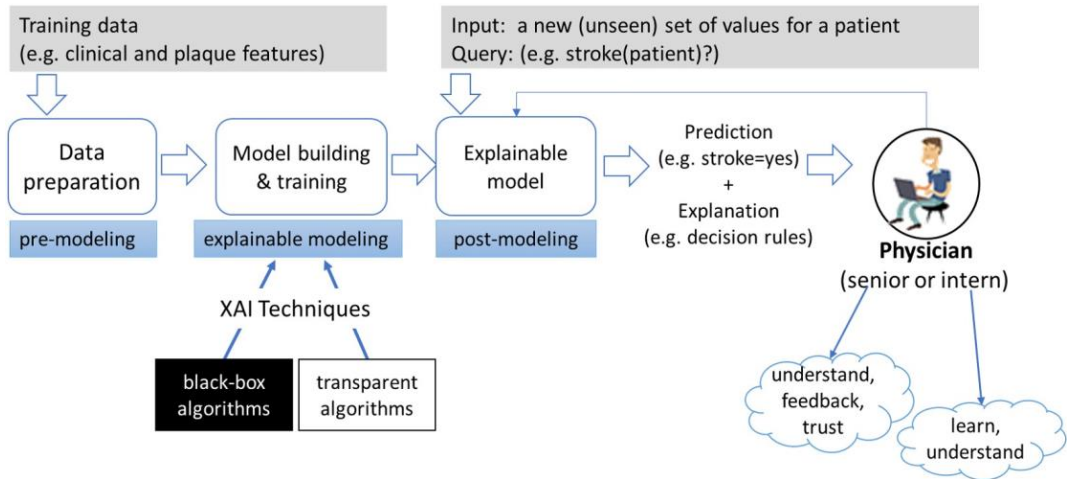

Fig. 3. Explainable AI(XAI) Design Concept. A medical use case example [10].

While a technique may belong to multiple categories for simplicity, each technique is assigned to a single category.

Common classification criteria include, (i) the notion of generality based on which the techniques are classified as *model-agnostic or model-specific*. (ii) Depending on the time the explanation is obtained, that is, during the learning process (training), by incorporating interpretability directly into the model structure, or, occurring after the learning process and treating the model as a black-box, techniques are classified as *post-hoc or ante-hoc* [9]. Further, (iii) explainability techniques can be classified as *global methods or local methods*, depending on the latitude/scope of the explanation, understanding the entire model's behavior, or understanding a single prediction. Additional criteria classify the result of the explanation technique, based on the *type* (feature statistics / model internals / data point) [20] and the *schema* (textual / graphical / visual) of the explanation, and based on the *questions* the explanation answers (what, why as attributive / why-not, what-if as contrastive / how as actionable) [72], [73]. Fig. 5 illustrates examples of graphical (a,c), visual (b,d) and textual (a,e) explanations.

In addition to the standard or typical XAI techniques, some works in the literature tried to address explainability with a *hybrid* approach that combines two or more different learning techniques. The integration of machine learning with symbolic AI methods is one example of these approaches, with recognized efforts in the areas of argumentation [10] - [74], knowledge graphs [75, 76], and fuzzy logic [77]. A recent work by MIT, DeepMind, and IBM [78] has also shown the power of combining connectionist techniques with symbolic reasoning.

### B. XAI Techniques

#### 1) Model Agnostic

Model-agnostic techniques treat a model as a black box aiming to provide an understanding of what knowledge has been acquired by the trained model. Some of these techniques can be applied to any machine learning model, e.g. when a framework/library (implementation) is available [79], [80], others require adjustments during the modeling stage, and others are agnostic for a particular category of models, e.g. deep learning architectures [81] or ensemble models [82]. These methods can be further grouped as follows:

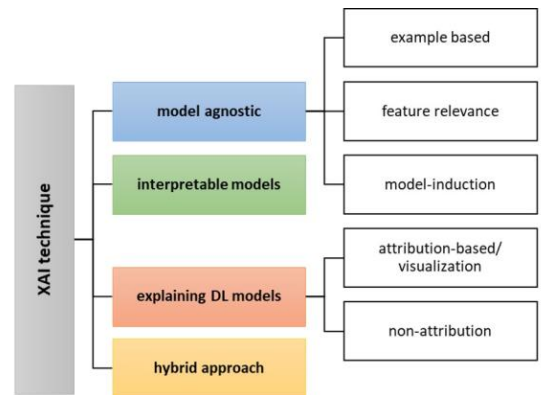

Fig. 4. XAI Techniques Taxonomy

1.1) Example-based:

Example-based techniques [83] select particular instances of the dataset to explain the predictions of a model.

1.2) Feature relevance

Feature relevance or importance or contribution techniques assign a score to the input features that indicates their relative importance to the prediction function of the model. In this way, these features offer an explanatory insight to the prediction.

- **Perturbation feature importance:** measures the importance of a feature by calculating the increase in a model's prediction error after permuting the feature [83]. A feature is considered important if the error is increased when shuffling its values.
- **Partial dependence plots (PDPs):** show the dependence between one or two input features and the target variable (output), by marginalizing over the values of all other features [83]. One-feature PDPs show the relationship between the target response and the feature (e.g. linear, non-linear). PDPs with two features show the interaction among these features. This technique requires that features in a plot are independent.

- **Shapley Additive Explanations (SHAP)** [84]: inspired by the coalitional game theory [77] and introduced to machine learning interpretation by Lundberg and Lee [71], builds on the assessment of combinations of features. Shapley value is the average marginal contribution of a feature value across all the possible combinations of features. SHAP implementations provide visualizations (plots) of features contribution to the prediction of a single observation (data instance), and summary visualizations as an overview of which features are most important for a model.

1.3) Model induction

Model induction techniques observe the behavior of an opaque trained model to infer a model that can be used to explain the predictions of the first.

- **Surrogate models**: are intrinsic interpretable models trained to approximate the predictions of a black-box model. They are usually trained on the inputs and predictions of the black-box model. Conclusions about the black-box model are drawn by interpreting the surrogate model. There are two kinds of surrogate models: global (approximate the predications of the black-box model as a whole) and local surrogate models (explain individual predictions of a black-box model, e.g. LIME [79]).
- **Knowledge distillation**: is a concept related to surrogate models, proposed in [85] and defined as the process of transferring knowledge from a large, slow, complex but accurate model to a smaller, faster, yet still accurate model. It is also known as mimic learning [51], a method where the complex model is a teacher/base model that trains a student/mimic model. Research works in this area use rule-based [86] or tree-based [85] surrogate models. Although these methods are considered model-agnostic, some techniques are more applicable to deep learning architectures such as distillation into Graphs [75].
- **Rule extraction**: is a recognized data mining technique widely used to reveal the hidden knowledge of opaque models such as deep neural networks (DNNs), support vector machines (SVMs), and ensemble models. Rule extraction is a model-agnostic approach, yet different algorithms have been proposed to accommodate the complexity of the various models. We provide some references about rule extraction from DNNs [87], [88], [89], from SVMs [90],[91] and from ensemble models [81].
- **ANCHORS** [80]: explains **individual** predictions by generating a decision rule that "anchors" the prediction sufficiently. A rule "anchors" a prediction locally if changes to the rest of the feature values do not affect the prediction. Therefore, anchors highlight the part of the instance that is sufficient for the classifier to make the prediction. Anchors' approach is based on reinforcement learning techniques in combination with a graph search algorithm [83].

2) Interpretable Models (by design)

Explainable AI can be achieved by designing or employing self-explanatory models that incorporate interpretability directly into their structure, and then explain their predictions in terms of the input features the model considers. Decision trees [92], rule-based models [50], additive modes [49], [93], sparse linear models [94], linear regression (Lasso) [95], fuzzy systems [96], [97] and bayesian models [98] are some examples. These models can provide both global and local explanations. Global explanations because of the transparent nature of the models, and local explanations by utilizing information of the model's parameters and structure (e.g. a path in a decision tree, a single rule, or the weight of a specific feature in a linear model).

TABLE IV
EXPLAINABILITY TECHNIQUES OVERVIEW

| Explainability method/ technique | Scope | Explanation Type |
|---|---|---|
| **Model Agnostic (*post-hoc*)** | | |
| Anchors [80] | L | Decision rules |
| Local Surrogate Models (LIME) [79] | L | Surrogate model |
| Shapley Values (SHAP) [84], [99] | L/G | Feature summary |
| Permutation Feature Importance [83] | G | Feature summary |
| Knowledge Distillation [51], [85], [100] | G | Surrogate model |
| Partial Dependence Plot [83] | G | Feature summary |
| Rule Extraction [81], [90], [101], [102] | G/L | Decision rules |
| RISE (for DNNs) [82] | L | heatmaps |
| Similar images (for DNNs) [103] | L | Image regions |
| **Interpretable Models (*model-specific, ante-hoc*)** | | |
| Linear models [94], [104] | G/L | Feature contribution |
| GAMs [49], GA$^2$Ms [93], [105] | G/L | Feature contribution, Features interaction |
| Decision Trees [106], [107] | G/L | Decision tress/ rules |
| Rule-based learning [50], [108] | G/L | Decision rules |
| Fuzzy-logic based [97] | G/L | Decision rules |
| Inductive Logic Programming [109] | G/L | Decision rules |
| **Explaining Deep Learning (DL) Models (*model-specific*)** | | |
| *Backpropagation-based attribution methods (**post-hoc**)* | | |
| Activation maximization [26], [110], [111] | L | Feature contribution |
| CAM and Grad-CAM [112],[113] | L | Feature contribution |
| DeepLIFT [114] | L | Feature contribution |
| Integrated Gradients [115] | L | Feature contribution |
| LRP [26], [66], [116], [117] | L | Feature contribution |
| Saliency maps [118], [119] | L | Feature contribution |
| *Perturbation-based attribution methods (**post-hoc**)* | | |
| Occlusion Sensitivity [22], [120] | L | Feature contribution |
| Sensitivity Analysis [121] | L | Feature contribution |
| Shapley value sampling [122] | L | Feature contribution |
| *Non-attribution methods (**post-hoc**)* | | |
| Concept vectors [123] | L | Concept contribution |
| Expert knowledge integration [124] | L | Rule-based |
| *Non-attribution methods (**ante-hoc**)* | | |
| Attention Mechanisms [125] - [126] | L | Feature contribution |
| Joint Training [127] - [128] | L | Feature contribution |

L: Local explanation. G: Global explanation.

3) Explaining Deep Learning Models

In a deep learning pipeline, a model learns representations from raw data and discovers patterns, i.e. mappings from representations to output (prediction). The learned deep representations are usually not human interpretable, therefore, many explanation techniques focus on understanding the representations captured by the neurons at intermediate layers of DNNs [129]. Likewise, in [26] the authors distinguish between DNN-interpretation, as the mapping of an abstract concept into a domain that humans can make sense of, and DNN-explanation, as the collection of features of the interpretable domain, that have contributed for a given example to produce a prediction. Combining the taxonomies suggested in [22], [62], [130] explainability techniques for deep learning architectures can be generally classified as illustrated in Fig. 4 in visualization/attribution-based and non-attribution (attention mechanisms and other techniques).

**a. Model agnostic** (DTX [131] example)

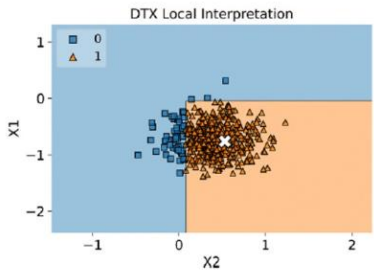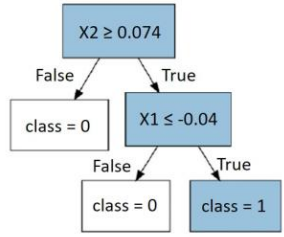

x1, x2: most important features around the neighborhood of x, the instance to be explained. class 0: negative class, class 1: positive class.

Examples of explanations of COVID-19 positive patients.

| ID | Decision Tree Explanation |
|---|---|
| 1 | EOS $\leq$ -0.51 **and** PLT $\leq$ 0.16 **and** CRP > -1.74 **and** EOS $\leq$ -0.61 **and** MPV > -0.82 **and** NEU $\leq$ -0.42 **and** MCHC $\leq$ 1.90. |
| 2 | CRP > -0.43 **and** EOS > 0.63 **and** AST $\leq$ -0.41 **and** UREA > -0.91 **and** MCV > 0.12 **and** CREAT > -0.88 **and** K+ > -0.52 |

EOS: eosinophils. PLT: Platelets. CRP: C-reactive protein. MPV: Mean platelet volume. NEU: Neutrophils. MCHC: MCH concentration. AST: Aspartate transaminase. UREA: Urea. MCV: Mean corpuscular volume. CREAT: Creatine. K+: Potassium.

**b. Explaining DL models** (multi-task CNN + attention mechanism)

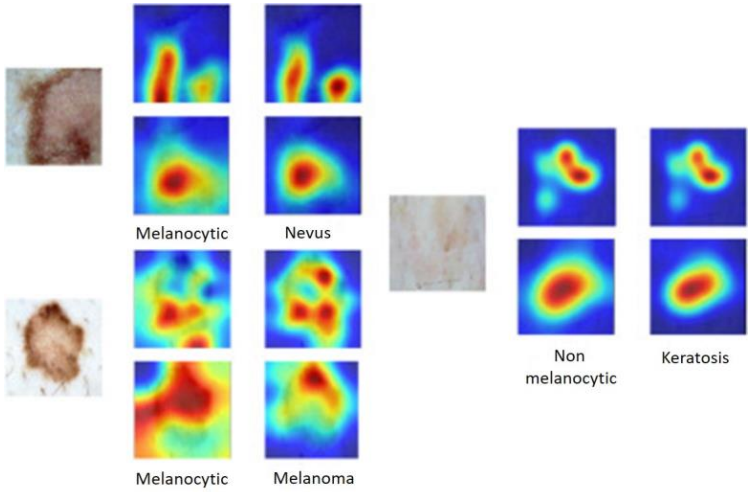

Correctly diagnosed skin lesions and the corresponding spatial attention maps [132].

**c. Interpretable by design** (a Bayesian Network example [133])

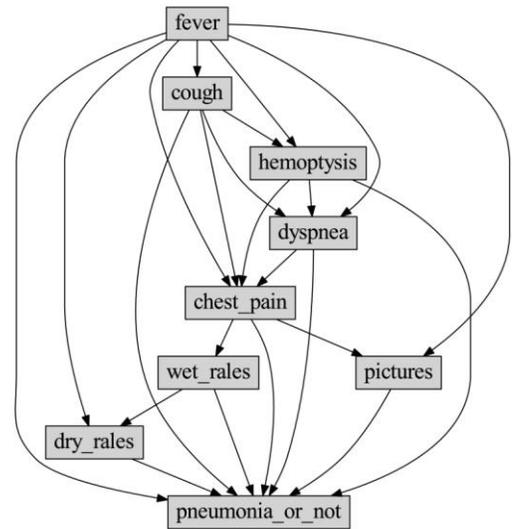

The network predicts the probability value of other nodes when fever occurs or not.

**d. Explaining DL models** (feature attribution methods examples) [134] – visual explanations for the prediction of a Non-Enhancing Tumor.

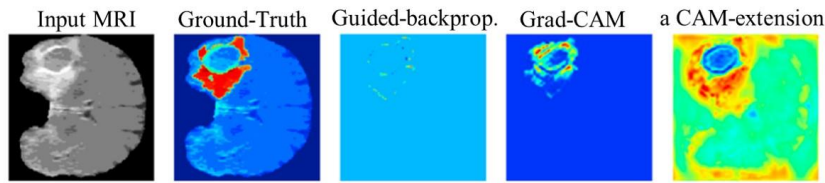

**e. A custom interpretable by design approach** – a Neutrosophic Clinical Decision-Making System [135].

| 1 | Input | Age=60, blood-pressure=143, cholesterol=290, and more (35 symptoms in total). | |
|---|---|---|---|
| 2 | Neutrosophication | Age(60)=($\mu_{young}$, $\mu_{middle-age}$, $\mu_{old}$)=(0,0.67,0), ($v_{young}$, $v_{middle-age}$, $v_{old}$)=(1,0.17,1), ($\lambda_{young}$, $\lambda_{middle-age}$, $\lambda_{old}$)=(1,0.4,1). Blood-pressure(143)=($\mu_{low}$, $\mu_{medium}$, $\mu_{high}$)=(0,0,0.8), ($v_{low}$, $v_{medium}$, $v_{high}$)=(1,1,0.4), ($\lambda_{low}$, $\lambda_{medium}$, $\lambda_{high}$)= (1,1,0.29). Cholesterol(290)= ($\mu_{low}$, $\mu_{medium}$, $\mu_{high}$)= (0,0,0.21), ($v_{low}$, $v_{medium}$, $v_{high}$)= (1,1,0.81), ($\lambda_{low}$, $\lambda_{medium}$, $\lambda_{high}$)=(1,1,0.8). | |
| 3 | Inference engine | Input | R60($\mu,v,\lambda$) | R64($\mu,v,\lambda$) |
| | | Age | middle (0.67, 0.17, 0.4 ) | middle (0.67, 0.17, 0.4) |
| | | Blood-pressure | high (0.8, 0.4, 0.29) | high (0.8, 0.4, 0.29) |
| | | Cholesterol | high (0.21, 0.81, 0.8) | high (0.21, 0.81, 0.8) |
| 4 | De-Neutrosophication | Mathematical calculations, example: <br> low=(0,0,1;0.3,1.2,1.2;0.2,0.9,0.9)=0+2(0)+1+0.3+2(1.2)+1.2+0.2+2(0.9)+0.9/12=0.605. <br> below medium=(0,1,2;0.2,1.5,2.5;0.1,1,1.7)=0+2(1)+2+0.2+2(1.5)+2.5+0.1+2(1)+1.7/12=1.125. <br> very high=(4,5,5;4.3,4.3,5,4.2,4.2,5)=4+2(5)+5+4.3+2(4.3)+5+4.2+2(4.2)+5/12=4.54. <br><br> **Output**: "The maximum risk value is of coronary artery disease" | |

Fig. 5. Examples of graphical (a,c), visual (b,d) and textual (a,e) explanation types.

*3.1) Visualization or attribution methods:*

Similar or identical methods are classified in the literature as visualization or attribution methods [62], [115]. *Visualization* methods associate the degree to which a DL model considers input features to a decision, by highlighting characteristics of an input that strongly influence the output [22]. *Attribution-based* methods provide a way of visualizing neural networks by performing a separate post-model analysis. Their goal is to determine the contribution of an input feature to the target neuron, by measuring the importance of the feature [115]. Frequently used visualization/ attribution methods can be further classified as *backpropagation-based* and *perturbation-based* [62].

- **Backpropagation-based methods:** visualize **features** of relevance based on the volume of gradient passed through network layers during network training. The attribution of the input features is computed with a single forward and backward pass through the network. Backpropagation-based algorithms include Activation Maximization [26], Gradients [115], Saliency Maps[136], LRP [137], CAM and Grad-CAM [112], Deep SHAP [138] and DeepLIFT [139].
- **Perturbation-based methods:** are considered the **simplest** way to analyze the effect of changing the input features on the output of an AI model. They visualize feature relevance by comparing network output between input and a modified copy of the input [22]. Occlusion sensitivity [22] and shapley value sampling [140] are two of the most commonly used perturbation-based techniques.

*3.2) Non-attribution methods:*

Non-attribution-based methods approach the problem of explainability by developing a specific methodology and validating it on a given problem, rather than performing a separate analysis using pre-existing methods. These include post-model methods, which may require specific changes to the model structure, and intrinsic methods that aim to improve the interpretability of internal representations within methods that are part of the deep learning architecture. These methods include attention-based mechanisms [141], concept vectors [123], joint training techniques [22], and other extended architectures.

## V. XAI APPLICATIONS IN THE MEDICAL DOMAIN

Despite the relatively recent emergence of the XAI research field, numerous solutions have been proposed, implemented, and tested on medical data. This section presents an overview of work in the literature on explainable AI/ML applications in the medical domain (subsections A and B), and a more detailed analysis of representative studies (domain expert in the loop) using explainability assessment metrics (subsection C).

### A. Medical XAI studies overview

As described in Section II the literature selection process resulted in 198 articles related to the scope of this review. We performed a high-level analysis of these articles considering the dimensions: (1) ML baseline model of the proposed solution, (2) XAI technique used, and (3) medical data. Fig. 6 illustrates the results of this analysis. The results confirmed that opaque (deep learning) models are the leading baseline method for XAI solutions for different medical tasks. seventy five percent (75%) of the articles examined proposed an explainable solution based on a deep learning architecture (Convolutional Neural Networks, Recurrent Neural Networks, and more). Eighteen percent (18%) of the papers proposed a solution based on an interpretable by design model (Decision tree, Linear Regression, Clustering algorithms, and more), and the remaining 7% of the papers were based on a multi-model solution. Regarding the usage of XAI techniques, and in relation to the taxonomy proposed in Section IV, 41% of the proposed solutions used a model-agnostic technique, 41% used a technique to explain the predictions of a DL model, 14% of the solutions used interpretable models, and the other 4% of the solutions were based on a hybrid approach. In the next paragraphs we discuss the techniques applied in each XAI category, based on the analysis in Fig. 6 (b) "XAI techniques usage" and provide more details and illustrations on representative works. Fig. 6 (c) illustrates the relation between the dimensions "medical data", "baseline model" and "XAI technique". The analysis results suggest that the most appropriate solution for the case of medical images and videos is the combination of opaque baseline models, for the prediction task, with explaining DL model (EDLM) techniques for the explanations. For the case of electronic medical records (EMR) the analysis results suggest using either an opaque baseline model with a model-agnostic XAI technique, or, an interpretable by design model and generate explanations using the model's insights. Similarly, for the case of omics and biosignals, the results suggest using a solution based on opaque baseline models combined with either a model-agnostic XAI technique or an EDLM technique. Finally, we observed that regarding the type of disease related to the medical data, the major disease groups identified are cancer and mental health (Alzheimer, Multiple Sclerosis, Parkinson, Schizophrenia) with 19% each, COVID-19/ Pneumonia with 14%, and cardiovascular with 12%.

### B. Medical XAI studies by XAI technique

*1) Model agnostic*

Shapley values (SHAP) is the technique most utilized in this category (~70%). It has been applied mainly on top of deep learning architectures [142] and ensemble models [143] and other interpretable models [144], [145]. While we expected LIME to be as popular as SHAP, only 22% of the studies in this category employed this technique. Other methods in this category include rule extraction [146],[147], and knowledge distillation [148].

In [149] the authors used both LIME and SHAP to generate explanations for an XGB-based model trained to predict 10-year overall survival of breast cancer patients, using data from the Netherlands Cancer Registry (46,284 patients and 31 features). Comparing the two techniques, the authors reported 95% consistency in explaining the contribution of the different features. They also argued that the regions of mismatched explanations allowed the identification of "turning points" in the features' values, where a feature goes from favoring one class (e.g. survived) to favoring the other class (e.g. deceased). In a similar study [150], a deep learning framework was

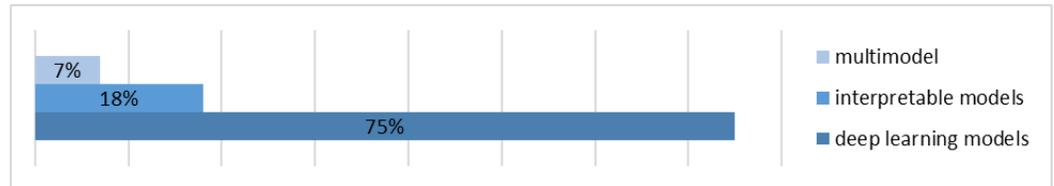
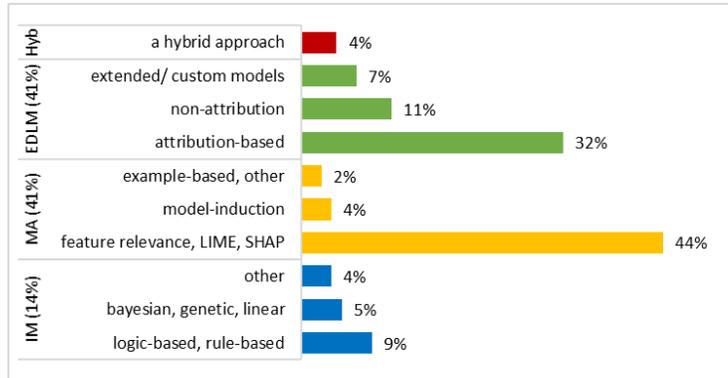
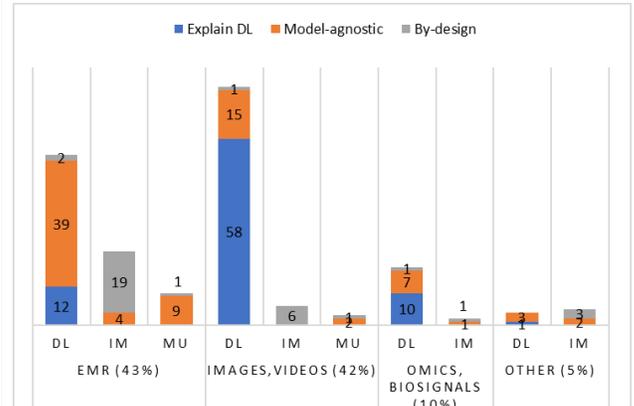

EDLM: Explaining DL models. MA: Model agnostic. IM: Interpretable models. Hyb: Hybrid.

DL: deep learning model. IM: interpretable model. MU: multi-model.

Fig. 6. Medical XAI studies overview. Analysis results.

developed to predict the brain age of a healthy cohort of subjects using their morphological features and DNN models. LIME and SHAP were used to explain the outcomes of the models and determine the contribution of each brain morphological descriptor to the final predicted age of each subject. The authors reported that the SHAP technique could provide more reliable explanations for the morphological aging mechanisms. A similar LIME approach is proposed in [131], known as Decision Tree-based Explainer (DTX), and defined as a model-agnostic, post hoc, perturbation-based, feature selector explainer. DTX generates a readable tree structure that provides classification rules, which reflect the local behavior of an ML model around the instance to be explained. Fig. 5 (a) illustrates how the DTX presents a visual output. The left side shows the noise set around the instance (x) that is going to be explained. The right side shows the tree structure generated by DTX for a local explanation. In the example in Fig. 6 (a), the explanation provided for why sample x is classified as class 1, is given by the path in the tree that leads to this outcome: $x_2 \geq 0.074$ and $x_1 \leq -0.04$.

*2) Interpretable models (by design)*

Rule-based learning [131],[151], network analysis based algorithms [98],[152], logic programming [153],[154],[10], patient similarity model (clustering) [155] and other transparent models were classified as "interpretable models" in the selected studies examined. A custom interpretable approach is presented in [135] and it refers to an explainable neutrosophic clinical decision-making system for cardiovascular diseases. Single-valued neutrosophic sets (SVNS) are used for decision-making. The system consists of five modules: neutrosophication, knowledge base, inference engine, de-neutrosophication, and explainability, with the explanation part being integrated with each module. The system takes 35 symptoms as input and determines the risk of eleven types of cardiovascular diseases. *Neutrosophication* is a process of converting crisp inputs in linguistic terms by determining the membership degree, indeterminacy degree, and falsity degree. The *knowledge base* contains 665 human-like judgments expressed as rules in the form of IF-THEN statements. The *inference engine* takes the linguistic value of the neutrosophication module and determines active rules out of all possible rules. The last part of the proposed decision-making process is *de-neutrosophication*. This part takes the firing strength of the active rules and maps these values on output membership functions of cardiovascular risks. The explanation of the proposed approach is straightforward as the system is transparent by design. An example is illustrated in Fig. 5 (e), but due to space constraints only a part of the system execution is shown here.

*3) Explaining Deep Learning models*

The majority of the techniques to explain a deep learning or opaque model, in the selected studies, were attribution-based. Other studies used an attention-based method or other custom approaches. In the next paragraphs, we provide more details on two examples to enlighten how the explanation is derived when using an attribution-based technique and a non-attribution one.

- **Attribution-based**: In [118] an extension of a CAM-based approach was proposed, to generate visual explanations for a 3D brain tumor segmentation network. The proposed methodology was applied on top of a 3D CNN model, based on the 3D Dilated Multi-Finer Network architecture. The model was trained on MRI images (ground truth MRI + four sequences for each subject) to predict Non-Enhancing Tumor, Edema, and Enhancing Tumor. The proposed XAI technique was based on a gradient-free method called Saliency Tubes [119]. During prediction, the activation maps of the convolutional layer to be visualized are extracted along with the predicted weights of the segmentation layer. Let Ak be the activation map of the lth layer and kth channel. A Hadamard product is obtained of each activation map of the lth layer with the predicted tensor of the segmentation layer followed by their weighted sum. Then, spline interpolation is used to increase the spatial information of the activation maps, and, min-max normalization is performed lastly to visualize the resultant saliency map. An example of these visual results is illustrated in Fig. 6 (d) (a CAM-extension), with Guided-backpropagation (GB) and Grad-CAM explanations to be part of the qualitative comparison performed in the study. The authors argued that their CAN-extension outperformed both the GB and the Grad-CAM techniques. Because of their dependency on the gradient information to generate the visual explanations, GB is not considered class-discriminative and Grad-CAM did not perform well on the overall coverage of the class of interest. Their gradient-free approach assisted in a more accurate visualization of the importance of each pixel with respect to each class, and in agreement with the domain knowledge of the domain experts. Computer-aided diagnostic systems (CADS) for the automatic analysis of medical images, which is mostly based on CNN models, are optimized to give the best classification performance, but they need to provide explainable outputs to the physicians. Attribution-based techniques and attention mechanisms are popular visualization methods used to improve the explainability of a CNN, by inspecting the features learned by the model. These techniques are applied in a post-hoc manner after the network is fully trained. Other methods try to improve the explainability and the performance of the model at the same time by training the network to jointly perform a set of related tasks. The next example is about a combination of multi-task CNN with visualization methods.

**- Non-attribution**: In [116] the proposed solution combines a multi-task CNN and attention mechanisms for the development of explainable CADS for skin cancer diagnosis. The system considers the inherent hierarchical structure of skin lesions to provide hierarchical decisions about a lesion at three levels: origin (melanocytic/non-melanocytic), degree of malignance (benign/malignant), and differential diagnosis (e.g., melanoma, basal cell carcinoma, benign keratosis), where each decision is conditioned on the previous one. The hierarchical diagnosis model is formed on three main modules: (i) image decoder (CNN based), which extracts discriminative features from raw images, (ii) image decoder (LSTM based), which performs the hierarchical classification; sequentially diagnosing the dermoscopy images, following the medical taxonomy, and, (iii) attention model that guides the model towards the most discriminative features and regions according to the previous output of the LSTM. The attention module's approach mimics the analysis performed by dermatologists while diagnosing skin lesions: inspect the lesion for localized criteria that give clues about its origin, followed by the detailed analysis of some of the identified criteria to perform the differential diagnosis. This analysis is reproduced by the incorporation of a spatial attention mechanism in the model. Spatial attention enforces the LSTM network to selectively focus on different parts of the skin lesion, considering the previously predicted hierarchical labels. The proposed model was developed using the ISIC 2017 and 2018 dermoscopy datasets with 2,750 and 11,527 images respectively. Different architectures and combinations of the various models were examined, with the best performing configurations achieving competitive results compared to other state-of-the-art solutions. An example of a spatial attention map is shown in Fig. 5 (b).

*4) Other-Hybrid*

An example of a twin system is the one presented in [133] **and called MulNet.** Inspired by the real diagnosis procedure of a human physician in pneumonia diagnosis, which involves the assessment of X-ray images, observable clinical features, and symptoms like fever, cough, and chest pain, the authors proposed the combination of deep learning and the Bayesian network to improve performance and interpretability. The proposed method integrates also multi-source data, electronic medical records (chest X-rays and textual reports of 35389 CAP (community-acquired pneumonia) cases) to predict pneumonia and medical knowledge for the Bayesian Network structure. First, a CNN DenseNet121 model was trained with chest X-rays images to predict the probability (0 or 1) of pneumonia. Then, a 7-dimensional vector was extracted from the report corresponding to the chest X-rays, indicating the presence or absence of the symptoms/indicators for pneumonia diagnosis (cough, hemoptysis, chest pain, fever, dyspnea, wet rales, dry rales). Label outputs from the CNN were derived with each 7-dimensional vector into 8-dimensional vectors. Finally, the Bayesian Network MulNet was trained using medical knowledge and the 8-dimensional vectors to diagnose pneumonia. Compared to SVM, RF, and DT, MulNet achieved the best classification performance of 87%. With regards to explainability, MulNet is considered an interpretable model by design. An example is shown in Fig. 5 (c). For any result of the diagnosis, the probability from root (result node) to leaf (factor nodes) can be analyzed to show the relationship between different factor nodes.

*C. Medical XAI studies, Explainability Evaluation*

In addition to the high-level analysis described in the previous paragraph (A. Medical XAI studies overview), we selected 10 representative articles, using the criterion of the domain expert participation in the design and/or evaluation of the proposed solution, to examine their explainability features and understand the effect of explanations on the issue of trust in Medical AI. Due to the lack of established methods for XAI solutions assessment, we identified common metrics from the selected studies to conduct this analysis. In particular, selected studies were examined using the following dimensions (see Table V):
- Medical AI information: ML baseline model, task, data.
- Explainability features: XAI technique, explanation type, local/ global, presentation.
- Evaluation goals/ metrics: accuracy, approach (questionnaire, survey, case study, model-expert comparison), experts (number), assessment remarks.

While Medical AI poses challenges to developers, medical professionals, and legislators, this analysis reflects mainly the challenges from the physician's perspective. Even though the current analysis is limited, in terms of assessment metrics, it is worth examining whether the evaluation results can meet the challenges discussed earlier.

TABLE V
MEDICAL XAI SELECTED STUDIES EVALUATION

| Medical AI Information | | | XAI Features | | | | Accuracy Assessment | | Human-centered Evaluation | | |
|---|---|---|---|---|---|---|---|---|---|---|---|
| Baseline | Data | Prediction | Tech. | Explanation | L/G | UI | Acc. | ALT | Approach | Ex. | Remarks |
| **Cancer** | | | | | | | | | | | |
| X-CFCMC [156] | medical images (N=5000) | colorectal cancer | by design | Image resp. to prediction | L | Visual | 93% | 94% | Quest. | 14 | + semantic explanation. + visualization influence trust. |
| BN CDSS, expert's input [70] | clinical guidelines (a), patient info. | laryngeal cancer therapy | by design | Observed evidence relevance | L | Visual, what-if | n/a | n/a | Quest. | 6 | + comprehension, justification of output. + simple visualizations. + what-if analysis. |
| CNN models [157] | endoscopic medical images (N=76) | Barret's esophagus cancer | multiple techs. (b) | Heatmaps (FC) | L | Visual | 84% | 79% | EMC | 5 | +model-human explanations agreement. + saliency attributes match best human's. |
| Case-based reasoning [158] | medical records (N=683) | breast cancer therapy | by design | Similar cases | L | Visual | 97.8% | 97.5% | User study | 11 | + visual UI. - inconsistent quantitative & qualitative similarities. |
| CNN [159] | MRI images (N=120) | Brain tumor | CAM | Heatmaps (FC) | L | Visual | 90% | 73% | Survey | 10 | + visual focus state. + acceptable model performance. + CAM best candidate XAI method. |
| | | | | | | | | | EMC | | tumor detection agreement: 99.3%. tumor type agreement: 93% |
| Fuzzy logic [160] | Experts knowledge | Hear-rate self-diagnosis | IM | Symptoms contribution (FC) | L | Visual | n/a | | Survey | 10 | 70% agreement: simple self-diagnosis rather than clinic visit. |
| RF-based [161] | EMR | Mortality risk | SHAP | Contributing factors (FC) | L | Visual | 94% | n/a | Focus group sessions | 21 | Explanation displays useful in assessing credibility, utility of prediction. Positive feedback about visual presentation and actions. |
| XGBoost-based [162] | Patient conditions, interview data, clinical decisions (N=18,840) | Laser surgery option | SHAP | Feature importance (G), one-vs-rest options FC (L) | L,G | Visual | 79% | 77% | Primary factors comparison | 9 | 92.7% agreement between the explainable model and physicians |
| CNN [163] | EHR to pathway images (N=6,675) | Sequences of conditions | Att-m | Clinical codes contribution (FC) | L | Visual | 85% (c) | n/a | EMC | 1 | Clinical codes mostly related and specific to predicted conditions. |
| VBridge [164] | EMR (N=1826) | Surgical complication | SHAP | Feature contribution (FC) | L, G | Visual | n/a | n/a | user study, interviews | 6 | + contextual explanations + what-if analysis + reduce mistakes risk |

L: Local explanation. G: Global explanation. UI: User Interface. CFCMC: Cumulative Fuzzy Class Membership Criterion, IM: Interpretable by design. At-m:attention-mechanism.RE: rule-extraction. CDSS: Clinical Decision Support System. BN: Bayesian Network. LRP: Layer-wise Relevance Propagation. EMR: Electronic Medical Records. ECM: expert-model comparison. FC: feature contribution. ALT: alternative model(s). (a) The TNM classification in laryngeal cancer management. (b) saliency, guided backpropagation, integrated gradients, input x gradients, DeepLIFT. (c) for predicting the sequence of predictions Birth Outcome → Perinatal Condition.

Candidate challenges to be examined, based on the available metrics:
(1) accuracy vs interpretability,
(2) quality of explanations: human-friendly explanations,
(3) human-in-the-loop: domain experts' active role in the design and evaluation of a medical XAI system.
(4) overall assessment, trustworthiness.
In the next paragraphs we discuss our findings.

*1) Accuracy vs Interpretability*
Explainability features should not alter the performance of the system in terms of prediction accuracy. This requirement is usually assessed with a comparison of the resulting model's accuracy with other baseline or state-of-the-art models. Looking at the columns "Accuracy assessment" in Table V we conclude that the majority of the solutions examined reported a better performance, compared to other models. We can conclude that the requirement of accuracy is generally met.

*2) Quality (human-friendly) explanations*
This is a broad requirement, it refers to all properties and features related to the ease of human understanding of the systems. Here we consider three properties of the explanation and one property of the system interface.

*2.1) Contrastive explanation:* why a prediction was made and why-not a different prediction, or why this prediction instead of other options. This property is related to the what-if analysis feature that some systems offer to the user through their interface. Based on the analysis in Table V (column UI) we can see that only 2 out of 10 studies offer this feature. Therefore, we can conclude that the requirement for contrastive explanation is hardly met.

*2.2) Selective/contextual explanation:* the lack of contextual information is one of the key challenges identified in [164] as this limits the use of decision support applications in clinical settings. Yet, this explainability feature was only examined in one particular study in which it was argued that contextual information helps clinicians understand explanations or "test their thoughts". We, therefore, conclude that the requirement for selective/contextual explanations is hardly met.

*2.3) Social interaction*: this means that explanations are part of social interaction between the explainer and the explainee. Medical professionals also expressed their thoughts of an ideal Medical XAI system as a "conversation partner". Looking at the column "UI" we understand that some systems offer an interactive interface, but only 3 out of 10 of the selected studies. Therefore, the requirement for social interaction is met at a minimum.

### 3) Human-in-the-loop

This is about the medical professionals' need to have a more active role in the development cycle of the systems, to be able to express their input during the design phase and provide their feedback on the resulting product, in terms of both the system and the functions it offers, as well as the recommendation, their explanation and the way these are presented to the end user. All studies in Table V had the domain expert involved in the explainability assessment. In three studies, [68], [70] and [164], the domain expert participated in the design/ development phase of the system. This finding is quite important, as it proves both the need to have the user in the loop and also, the physician's readiness to be part of this process. Obviously, the requirement for human-in-the-loop is met, in this small number of studies, however, compared to the total number of studies included in this review the requirement is 30% met.

### 4) Overall assessment, trustworthiness

Although explainability is used as a means to overcome the lack of trustworthiness in Medical AI, very few studies clearly discuss their findings in relation to trust. In [159] doctors agreed that "the system is trustworthy in terms of diagnosis of brain tumors". They also reported that the performance of the model is acceptable, and that the XAI view of the system enabled them to make decisions faster. Explainability can promote trust, support the human-AI interaction and improve the decision-making process. Similarly in [156], and based on the pathologists responses in the questionnaire given, "the semantic explanation" and "the visualization of the training image responsible for the prediction" increased their trust. Finally, the otolaryngologists participated in the evaluation study in [70], for the assessment of the proposed approach to generate trustworthy and justifiable results, recommended its usage within clinical routine.

### D. Revised Medical XAI concept

In Section IV we began our discussion about XAI techniques using the standard machine learning pipeline. Considering the physician's feedback, and findings from this review, we suggest a revised design concept, the "Medical XAI Design Concept". As illustrated in Fig. 7 the physician (domain expert) has a more active role in the development of AI systems, by providing the medical expertise at the beginning of the process, evaluating the results, and providing feedback on the final output and functionality of the system. Finally, the explanation can help a physician understand the recommendations provided by the system, and decide to trust the system or not, provide feedback or suggest changes for its improvement, learn and discover from the findings, and be able to take a more informed decision (accountability).

## VI. CONCLUSIONS

Explainability is a multidimensional concept that is widely discussed in the literature, with 30% of the relevant work located in the field of medicine. Despite the lack of a universal definition, researchers agree that the ultimate goal is to make AI systems produce results that can be better understood by humans in a natural way. Because of ethical, legal, technical issues and the need to develop fair and safe systems that we can trust, explainability is considered essential for the successful deployment of AI solutions in practice. In the medical domain, explainability is of particular importance because of the significance that medical decisions can have on the lives of people. In many cases decisions in medical care are taken in consultation between peers and experts. For AI systems to integrate naturally within such consultations they will need to be highly explainable with meaningful and useful explanations. It is also argued that "omitting explainability in clinical decision support systems poses a threat to core ethical values in medicine and may have detrimental consequences for individuals and public health".

In this study, we reviewed explainable AI/ML solutions applied in the medical domain to report recent developments (concentrating on work between 2019 to October 2022) and identify areas that further research is needed. We based our review on a set of key questions for which we summarize here the findings revealed by the review:

(i) "*Which XAI techniques have been used and for which medical use case?*". Model-agnostic techniques (mainly SHAP) and feature attribution techniques on top of deep learning architectures were mostly applied in the studies examined, to provide explanations for a medical diagnosis, a specific diagnostic sub-task, or therapeutic decision support.

(ii) "*Is there a connection between the XAI techniques and the medical use case or the medical data the solution has been applied to?*". We did not identify any pattern between the XAI technique and the medical task of the various solutions. The selection of the XAI technique depends on the AI/ML baseline model, which depends on the prediction task and the type of data.

(iii) "*Is the explanation useful, has it been evaluated by the physicians?*". To answer this question we identified a number of studies that reported some kind of explanation assessment. While very few works have reported the physician's involvement in the evaluation of the explanation, they considered the explanations important in understanding the recommendations of the system and increasing their trust.

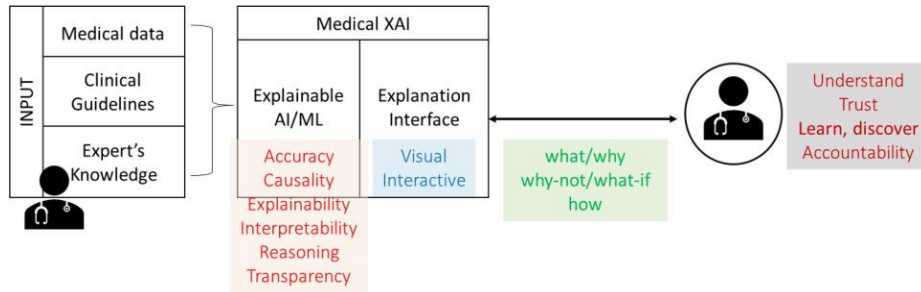

Fig. 7. Proposed Medical XAI Design Framework (with features and properties distributed in the pipeline).

(iv) *"Can we apply these solutions in clinical practice?"*. No study has reported the integration of these solutions in clinical practice, though, some prototypes were made available to the end-users for evaluation purposes. To the physician, the explanation is just another piece of information that needs to be examined and evaluated. Unless there is a way for a systematic assessment of these systems, that will safeguard the accuracy and correctness of their function and results, it would be difficult to apply these solutions in clinical practice.

(v) *"To what extent are the explainability challenges addressed by the proposed solutions and which areas need further development?"* . To answer this question we examined the challenges relating to the explainability assessment metrics discussed in Section V, paragraph C. Our findings suggest that these challenges were met to a minimum. However, this exercise was not based on a formal evaluation approach, as the field of explainability assessment is still in its infacy. Nevertheless, the result is an indication that more work is needed to meet the different challenges and measure the effectiveness of the explainability results.

Enabling explanations within an AI/ML model is a challenging task. While various solutions have been proposed and implemented they are currently focused on deriving features or facts from the model to justify their prediction. Other factors need to be considered, like the recipient of the explanation and its intended use (e.g. debug, verify, understand), or the need for contextual explanations depending on the user role (developer, physician, regulator) or the level of expertise (e.g. junior or senior practitioner). Approaching explainability from the point of the user needs is a pattern in the literature, yet, application and regulatory requirements should also be considered. More effort is required to understand the explainability requirements, from a human, legal, medical and technical perspective, how to design and implement accordingly these systems, how to measure and evaluate the quality of explanations and how such methods can be standardized in the field of medicine. Our findings indicate that close collaboration between medical and AI experts, could guide the development of dedicated frameworks for the design and implementation of XAI systems in medicine, and provide appropriate tools for their systematic evaluation. Our findings also suggest that the way the explanation is presented, in terms of user interface and user experience, is important for the efficient user-AI interaction and special work is needed towards this direction.

In Medicine, where the main group of users is that of the medical practitioners, the explanations need to be well-informed in terms of the underlying medical theory, official guidelines and best practices. This is a complex problem that goes beyond today's predominant methods of explaining a model generated purely by some ML method. In this task, hybrid approaches of neuro symbolic integration could help provide models that encompass a richer form of explanations that are cognitively compatible with their expert users.

---

[i] Selected articles list of references: https://refworks.proquest.com/public-share/9pXanH5MX8Z8rqDN06pA9kThaQm7gYlmEdKXDF53JEyQ